\documentclass{article}

\PassOptionsToPackage{numbers, compress}{natbib}

\usepackage[final]{DSA_2025}

\usepackage{amsmath}
\usepackage{fancyhdr}
\usepackage{lastpage}
\usepackage{tcolorbox}
\tcbuselibrary{skins}

\usepackage{xcolor} 
\usepackage{enumitem} 
\usepackage{boxedminipage}
\usepackage{graphicx}
\usepackage[utf8]{inputenc} 
\usepackage[T1]{fontenc}    
\usepackage{hyperref}       
\usepackage{url}            
\usepackage{booktabs}       
\usepackage{amsfonts}       
\usepackage{nicefrac}       
\usepackage{microtype}      
\usepackage{xcolor}         

\title{Temporal Analysis of Climate Policy Discourse: Insights from
Dynamic Embedded Topic Modeling}

\author{%
  Rafiu Adekoya Badekale\thanks{Code: The code for this paper can be found at  https://github.com/AdeTheBade/TACPD.git} \\
  Hamoye Foundation\\
  Delaware, United States \\
  \texttt{ade.badekale@hamoye.org} \\
  \And
  Adewale Akinfaderin \\
  The George Washington University \\
  Washington, D.C., United States \\
  \texttt{wale@hamoye.org} \\
}

\begin{document}

\maketitle

\begin{abstract}
Understanding how policy language evolves over time is critical for assessing global responses to complex challenges such as climate change. Temporal analysis helps stakeholders, including policymakers and researchers, to evaluate past priorities, identify emerging themes, design governance strategies, and develop mitigation measures. Traditional approaches, such as manual thematic coding, are time-consuming and limited in capturing the complex, interconnected nature of global policy discourse. With the increasing relevance of unsupervised machine learning, these limitations can be addressed, particularly under high-volume, complex, and high-dimensional data conditions. In this work, we explore a novel approach that applies the dynamic embedded topic model (DETM) to analyze the evolution of global climate policy discourse. A probabilistic model designed to capture the temporal dynamics of topics over time. We collected a corpus of United Nations Framework Convention on Climate Change (UNFCCC) policy decisions from 1995 to 2023, excluding 2020 due to the postponement of COP26 as a result of the COVID-19 pandemic. The model reveals shifts from early emphases on greenhouse gases and international conventions to recent focuses on implementation, technical collaboration, capacity building, finance, and global agreements. Section 3 presents the modeling pipeline, including preprocessing, model training, and visualization of temporal word distributions. Our results show that DETM is a scalable and effective tool for analyzing the evolution of global policy discourse. Section 4 discusses the implications of these findings and we concluded with future directions and refinements to extend this approach to other policy domains.
\end{abstract}

\section{Introduction}
Climate change remains one of the most pressing global challenges of the 21st century, with far-reaching environmental, economic, and social implications \citep{ipcc2022}. The United Nations Framework Convention on Climate Change (UNFCCC), established in 1992, serves as the institutional backbone for global climate governance \citep{yamineva2024}. Its annual Conference of the Parties (COP) meetings provide a platform for negotiating and implementing climate policies. Over the past three decades, the discourse within UNFCCC decision documents has evolved significantly. This shift reflects how nations strive to meet sustainability goals and honor commitments under international agreements such as the Paris Agreement \citep{backstrand2017non}. As climate concerns continue to grow in complexity, the corresponding policy discourse has also expanded in scale, depth, and multidimensionality.

Understanding how climate-related policy themes change over time is critical for effective decision-making. It helps policymakers and researchers identify gaps, anticipate future directions, and address dynamic global needs \cite{Schmidtn}. Traditional methods, such as qualitative reviews or simple trend analyses, fall short for several reasons. These approaches are labor-intensive, requiring extensive manual effort \citep{hsu2021}. They struggle to keep pace with the growing volume of policy documents. Moreover, they often fail to capture the interconnected dynamics of climate issues, such as the interplay between energy, technology, finance, and international cooperation \citep{keohane2016cooperation}. To address these limitations, our main contributions are summarized as follows,

\begin{itemize}
    \item Demonstrate the applicability of advanced topic modeling to real-world policy analysis.
    \item Trace the evolution of global climate policy discourse.
    \item Identify major thematic shifts and emerging priorities.
\end{itemize}

Recent advances in natural language processing (NLP) enable large-scale text analysis with minimal supervision \citep{blei2012probabilistic}. Topic modeling, a subfield of NLP, reveals latent thematic patterns without relying on predefined categories. This study applies the dynamic embedded topic model (DETM) to a corpus of UNFCCC policy decisions from 1995 to 2023, excluding 2020 due to the postponement of COP26 as a result of the COVID-19 pandemic. DETM integrates word embeddings and dynamic structures to model time-dependent changes in topic prevalence and content \citep{dieng2020}. This capability is particularly valuable for climate policy research, where understanding historical shifts can provide data-driven insights for policymakers. Our approach aims to enhance the scalability, transparency, and relevance of analyzing policy discourse, particularly within the context of global climate negotiations under the UNFCCC.

\section{Related Work}

Our novel approach builds on climate policy discourse, topic models, and dynamic embedded topic models.

The analysis of climate policy discourse has long been a subject of interest in environmental governance, sustainability studies, and political science. Scholars employed qualitative and quantitative methods to study how climate issues are framed, negotiated, and institutionalized over time \citep{keohane2016cooperation, victor2011global}. Early studies analyzed international agreements such as the Kyoto Protocol, focusing on legal frameworks, implementation challenges, and impacts on global emissions. For instance, Aichele and Felbermayr \citep{aichele2013} evaluated the effectiveness of the Kyoto Protocol in driving emission reductions, highlighting the role of procedural mechanisms in early climate governance. Recent studies explored the post-Paris Agreement era. They emphasized the shift toward broader cooperation mechanisms, such as climate finance and capacity building, as critical components of global climate strategies \citep{dimitrov2019, bowman2024, ullah2025}. These traditional approaches have largely relied on content analysis, expert interviews, and manual coding of policy documents. Although insightful, they are labor-intensive and limited in scalability, particularly when working with large datasets \citep{hsu2021}. These limitations are particularly evident when analyzing decades of policy discourse, where discussions are multidimensional and evolve in nonlinear ways.

With the rise of computational social science and NLP, researchers have increasingly adopted automated approaches for analyzing large-scale policy corpora. Topic models are useful tools for the statistical analysis of document collections \cite{yang2011}. They have been applied across various fields, including marketing, sociology, political science, and the digital humanities. One of the most common topic models is latent Dirichlet allocation (LDA) \cite{blei2003}. A probabilistic model that represents each topic as a distribution over words and each document as a mixture of the topics. LDA remains one of the most widely used topic modeling methods. In the context of climate discourse, Farrell \cite{farrell2016} explored framing strategies, while Mildenberger et al. \cite{mildenberger2017} investigated partisan divides in U.S. climate rhetoric using LDA. However, LDA and similar static models assume a fixed topic distribution over time,  which limits their ability to capture temporal dynamics in policy narratives.

To address this temporal limitation, researchers developed the dynamic topic model (DTM) \cite{blei2006}. The earliest and most widely used version is the dynamic latent Dirichlet allocation (D-LDA) model. DTM incorporates time by modeling the evolution of topic distributions across sequential batches of document \citep{yeh2016}. It has been used to study policy evolution in various domains, including environmental regulation and public health. For instance, Wang et al. \citep{wang2024} applied a DTM to China’s energy policy documents. The model revealed a shift in policy focus over time from infrastructure development and standardization management to new energy development and modernization of the energy system. Despite these advancements, DTM relies on linear transitions and bag-of-words representations, which restrict its semantic depth. These limitations are particularly evident when analyzing long timeframes.

To improve semantic coherence, more recent models have incorporated word embeddings. The embedded topic model (ETM) \cite{dieng2019} uses embeddings and amortized inference to produce more interpretable topics. Building on this, DETM combines word embeddings with recurrent neural networks to model topic evolution over time. The model uses a variational autoencoder to capture temporal patterns and generate more semantically rich topics \cite{dieng2020}. Although topic models have shown promising results in domains such as news, historical records, and scientific publications, they remain underexplored in environmental policy research.

This research extends the application of DETM to the domain of climate governance by analyzing UNFCCC decision documents spanning nearly three decades. Previous studies, such as Schmidt et al. \cite{schmidt2013}, examined framing patterns in climate negotiations using discourse analysis. More recent work has applied machine learning techniques to monitor public sentiment \citep{nurlanuly2025} or track climate misinformation \citep{robillard2024}. However, to the best of our knowledge, no study has conducted a temporal, data-driven analysis of official climate policy texts using advanced probabilistic models. By applying DETM to this global corpus, this study helps fill that gap. It offers a scalable framework to uncover latent thematic shifts and track evolving policy priorities. This positions the study at the intersection of climate policy analysis, temporal topic modeling, and machine learning-based document understanding.

\section{Methodology}

\subsection{Data Collection and Preprocessing}
We collected decision documents from the UNFCCC website, spanning the years 1995 to 2023, excluding 2020 due to the postponement of COP26 as a result of the COVID-19 pandemic. This resulted in 28 distinct time steps (1995–2019, 2021–2023), each representing a year of COP proceedings. This period covers significant milestones in climate governance, including the adoption of the Kyoto Protocol (1997), the Paris Agreement (2015), and recent efforts to operationalize global climate strategies. We extracted the texts from each document and grouped them by year in a CSV file to reflect temporal structure. To ensure data accuracy, we conducted random checks on the CSV file. We manually verified the content, confirming the reliability of our collation process.

\begin{figure}[ht]
  \centering
  \includegraphics[width=0.9\textwidth]{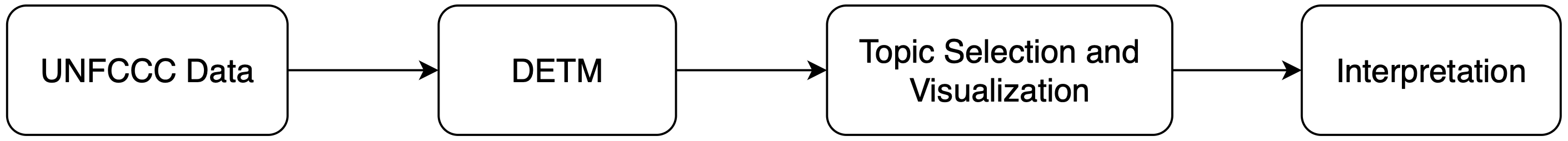}
  \caption{Flow diagram of study}
\end{figure}

Preprocessing was performed to prepare the texts for DETM training. Using the DETM \cite{dieng2020} preprocessing script, we applied standard NLP techniques. We split the texts into paragraphs to ensure finer granularity, as paragraph-level analysis better captures thematic coherence. We filtered out stopwords, punctuation, and low-frequency terms. To balance vocabulary size and informativeness, we set a minimum document frequency threshold (min\_df=100), retaining only terms that appeared in at least 100 documents. We allocated 85\% of the documents for training, 10\% for testing, and 5\% for validation. This process resulted in a vocabulary of 960 unique terms. The preprocessed data, including the vocabulary, tokenized texts, and timestamps, were saved for model training. Table \ref{tab:dataset_stats} summarizes the characteristics of the dataset.

\begin{table}[ht]
\centering
\caption{Summary statistics of the dataset under study.}
\label{tab:dataset_stats}
\small
\begin{tabular}{lp{0.4in}p{0.4in}p{0.4in}p{0.3in}p{0.4in}}
\hline
Dataset & Train Docs & Val Docs & Test Docs & Time-stamps & Vocab. \\ \hline
UNFCCC \rule{0pt}{1.5em} & 196,290 & 11,563 & 23,097 & 28 & 960 \\ \hline %
\end{tabular}
\end{table}

\subsection{DETM Training}
We implemented the DETM training using a Python script (main.py) from the DETM \cite{dieng2020} repository, adapting it to our dataset. Dependencies were managed through supporting scripts (detm.py, utils.py, and data.py). We configured the model's hyperparameters experimentally. The number of topics (num\_topics) was set to 5, the hidden state size for the topic proportion network (t\_hidden\_size) to 800, and the embedding size (rho\_size) to 300, to align with pre-trained word embeddings. The model was trained for five epochs, which we found sufficient for convergence based on validation loss. The training process generated a beta matrix, representing the topic-word distributions across the 28 time steps. This matrix was saved for subsequent analysis.

\subsection{Topic Selection and Visualization}
We selected topic 2 based on its relevance to Carbon Capture and Storage (CCS), a key area in climate mitigation. The selection was guided by keyword matches using a predefined set of CCS-related terms (e.g., “carbon,” “capture,” “storage,” “CCS”). To analyze the evolution of climate policy discourse, we developed a visualization python script (plot\_topic\_evolution.py) that extracts and displays the top words for the selected topic across specific time points, based on the beta matrix.

\begin{figure*}[ht]
    \centering
    \setlength{\fboxsep}{5pt} 
    \setlength{\fboxrule}{2pt} 
    \fbox{\includegraphics[width=0.9\linewidth]{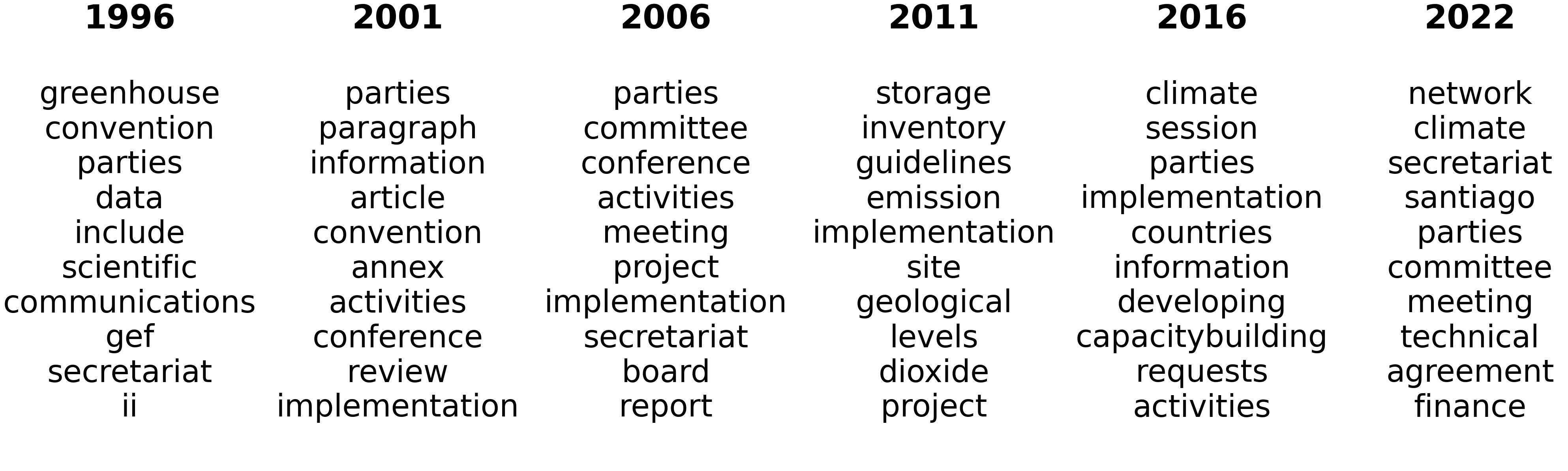}}
    \caption{Temporal evolution of topic 2 ("Climate Policy Mechanisms and Mitigation Strategies")}
    \label{fig:topic2_evolution}
\end{figure*}
\begin{figure*}[ht]
    \centering
    \includegraphics[width=0.9\linewidth]{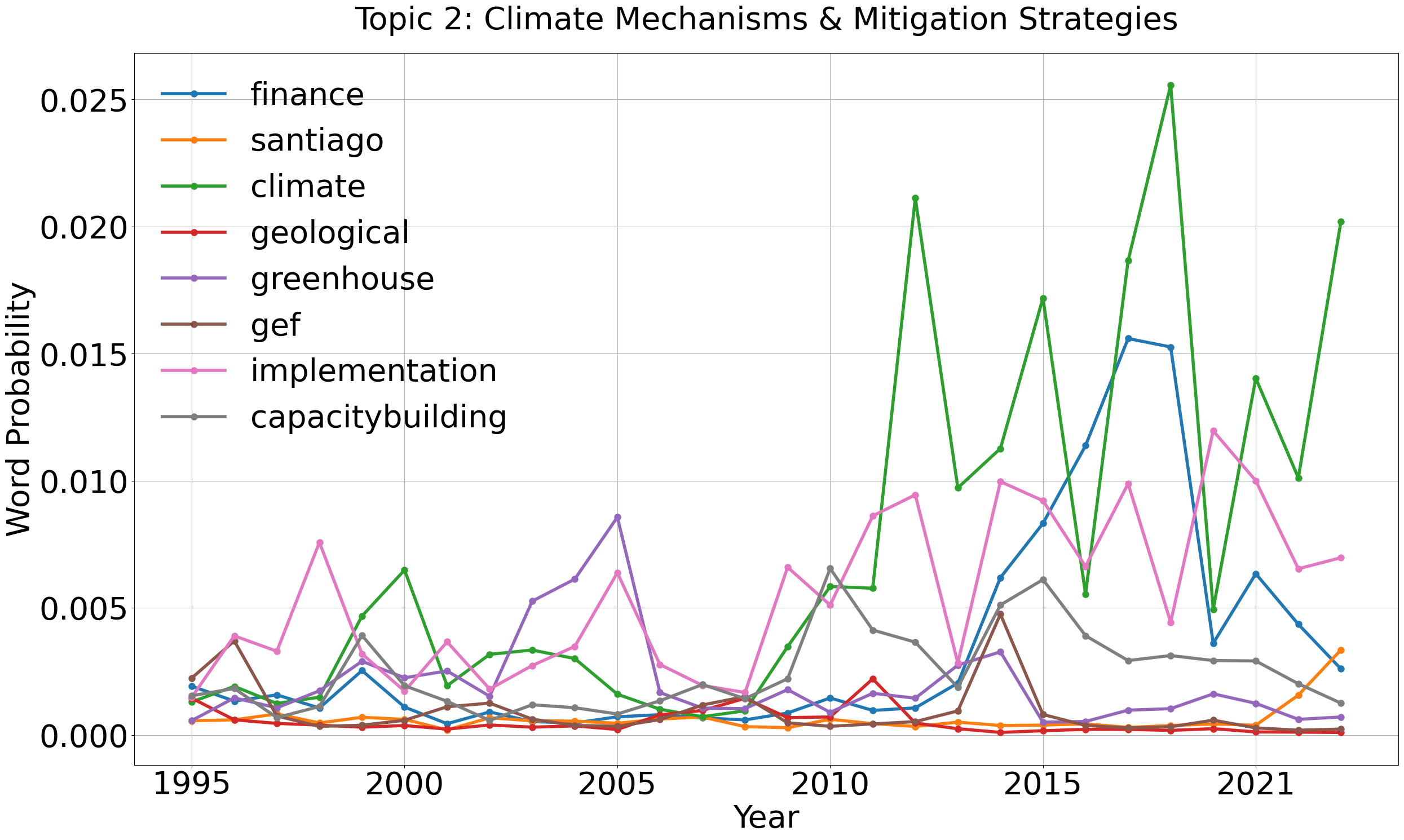}
    \caption{Evolution of query words in topic 2, showing the probability trends of key terms}
    \label{fig:query}
\end{figure*}

For visualization, we chose six-year intervals to capture long-term trends, as shown in Figure~\ref{fig:topic2_evolution}. The words were plotted in a subplot format, with each subplot displaying the top 10 words for a given year. This provides a visual representation of topic evolution over time. Figure \ref{fig:query} shows the probability trends of key terms in topic 2. To provide a comprehensive view of the discourse structure, we extended this visualization approach to the remaining four topics. Their top 10 words over time are presented in a format similar to Figure~\ref{fig:topic2_evolution}, included in the appendix. This approach offers a clear, interpretable view of how policy themes, such as those related to CCS or climate finance, have shifted over time.

\subsection{Interpretation of Policy Evolution}
We interpreted the results through a policy lens by aligning the dominant terms and their temporal peaks with historical events, negotiation themes, and major UNFCCC agreements. This interpretation adds context to the DETM output and provides actionable insights for policymakers seeking to understand how global climate policy discourse has evolved.

\section{Experimental Results and Discussion}

\subsection{Evolution of Climate Policy Discourse}
\textbf{1995–2000: Foundational Frameworks}

The early years were defined by foundational terms such as "convention", "parties", and "communications", reflecting initial international cooperation and treaty formation, notably the Kyoto Protocol (adopted 1997, entered into force 2005) \citep{victor2011global}. Terms such as "greenhouse", "data" and "scientific" indicate a priority on understanding greenhouse gas emissions through data collection and scientific communication. References to "gef" (Global Environment Facility), and "secretariat" reflect institutional groundwork to support the UNFCCC. The appearance of "ii" points to the roles of Annex II countries, indicating early debates around differentiation and responsibility. This period centered on setting agendas and building procedural frameworks for global climate governance.

\textbf{2001–2010: Operationalizing Agreements}

This decade signaled a shift toward implementation, with high-frequency terms including "project", "activities", "committee", "review", and "report". These reflect operationalization of earlier agreements and enhancing institutional coordination \citep{keohane2016cooperation}. Terms like "meeting" and "report" highlight the role of COP meetings in monitoring progress and global negotiation. This transition from normative goal-setting to practical execution mirrors the maturation of climate governance, as countries began implementing early commitments.

\textbf{2011–2016: Technical Mitigation Strategies}

The discourse became more technical. Terms like "storage," "geological," and "dioxide" signals a focus on Carbon Capture and Storage (CCS) as a mitigation strategy. This aligns with global efforts to develop CCS technologies in the late 2000s and early 2010s \citep{ansari2019global}. "Guidelines" and "inventory" suggest the development of standards for CCS deployment and emissions tracking. By 2015, with the adoption of the Paris Agreement, terms like "developing", "capacity building", and "finance" gained prominence, reflecting increased attention on equity, technical assistance, and support for developing nations through financial and technical mechanisms \citep{backstrand2017non}.

\textbf{2017–2023: Global Collaboration and Support Mechanisms}

Recent years emphasized implementation through global collaboration, reflected in terms like "network", "technical", "agreement", and "finance". The term "santiago" refers to the Santiago Network, under the Warsaw International Mechanism (WIM) for loss and damage, indicating operationalization of support mechanisms \citep{antypas2024saving}. "Technical" and "finance" reflect the emphasis on technical collaboration and funding as key pillars of post-Paris climate governance. The continued presence of "secretariat", "committee", and "meeting" indicates ongoing institutional engagement.

\bigskip
These trends mirror the maturation of climate policy, from early institutional design to implementation-focused, globally inclusive strategies under frameworks like the Paris Agreement.

\subsection{Temporal Word Distributions}
DETM’s ability to model latent topic proportions over time enabled us to capture transitions in discourse emphases. Figure \ref{fig:query} visualizes key query word distributions for topic 2 across selected years,  further confirming the previously defined trends. Terms like "greenhouse" and "GEF" emerged in the early years. "Implementation" and "climate" showed sustained relevance over the years. "Geological" peaked around 2010, reflecting the CCS focus. "Finance," and "capacity building" emerged strongly in the last decade, signaling a policy focus on the post-Paris emphasis on climate finance and technical collaboration.

\subsection{DETM as a Lens for Policy Dynamics}

Understanding how climate discourse has changed provides a roadmap for anticipating future developments. These findings demonstrate DETM’s ability to capture policy shifts over time, unlike static content analysis. This capacity makes DETM particularly valuable for policymakers and researchers aiming to evaluate past priorities, identify emerging themes, and align strategies with global discourse. For example, policymakers can use DETM to identify when specific issues, such as CCS or climate finance, entered the discourse and how they evolved. This can inform retrospective evaluations and guide the timing of new initiatives. Delegates preparing for COP negotiations can also leverage DETM insights to align proposals with historical discourse trends or identify emerging topics gaining momentum. This can ensure better framing and positioning during discussions.

\subsection{Toward a Broader Research Agenda}

This study on UNFCCC decision documents highlights DETM’s potential as a scalable tool for policy analysis. It reveals long-term temporal dynamics that can inform climate governance and beyond. DETM can also be applied to fields such as sports, health, or education. For example, it could be used to analyze shifts in public health priorities in World Health Organization (WHO) reports. However, DETM’s performance is sensitive to corpus quality and preprocessing. Important contextual terms, especially in technical or legal language, may be lost due to tokenization or frequency thresholds. Its unsupervised nature requires domain expertise to interpret ambiguous terms, such as "ii" (1996, Annex II countries) or "santiago" (2022, Santiago Network). Additionally, DETM focuses on word distributions and does not capture sentiment or argumentation structure, which are often crucial in policy analysis. To address these limitations, future work could integrate DETM with BERTopic to enhance semantic understanding \citep{grootendorst2020}. Incorporating sentiment analysis and involving stakeholders to define priority themes would further improve interpretability and ensure alignment with specific policy needs. These improvements would strengthen the model’s relevance and increase its practical utility across diverse policy contexts.

\section{Conclusion}

This study demonstrates the effectiveness of DETM in analyzing nearly three decades of UNFCCC decision documents. The model reveals the evolving landscape of climate policy discourse from foundational agreements and early concerns about emissions to more recent focuses on capacity building, finance, and collaborative implementation. Our findings highlight the growing complexity and diversity of climate discourse, emphasizing the need for adaptive, evidence-based policy frameworks. By offering a data-driven perspective, DETM enables researchers and policymakers to monitor shifting priorities, detect emerging narratives, and anticipate future directions in global climate strategy. This becomes particularly valuable in an era where the volume and complexity of policy-relevant texts are rapidly increasing. Leveraging machine learning in this way holds promise to strengthen evidence-based decision-making, enhance transparency, and support more effective international cooperation in the fight against climate change.

\begin{ack}
This work is fully funded by Hamoye Foundation, a 501(c)(3) organization.
\end{ack}



\newpage

\section*{Appendix}
 

\appendix
This section provides additional details to support the main findings of the paper. The appendices include the temporal evolution of additional topics (Appendix A) and a glossary of acronyms used in this study (Appendix B). The code for the topic modeling, along with the UNFCCC policy decision dataset, is available at [GitHub link: https://github.com/AdeTheBade/TACPD.git] for reproducibility.

\section{Temporal Evolution of Additional Topics}
The following figures present the temporal evolution of the top 10 words for the four additional topics (excluding topic 2, which is detailed in the main text).

\begin{figure}[ht]
    \centering
    \setlength{\fboxsep}{5pt} 
    \setlength{\fboxrule}{2pt} 
    \fbox{\includegraphics[width=0.9\linewidth]{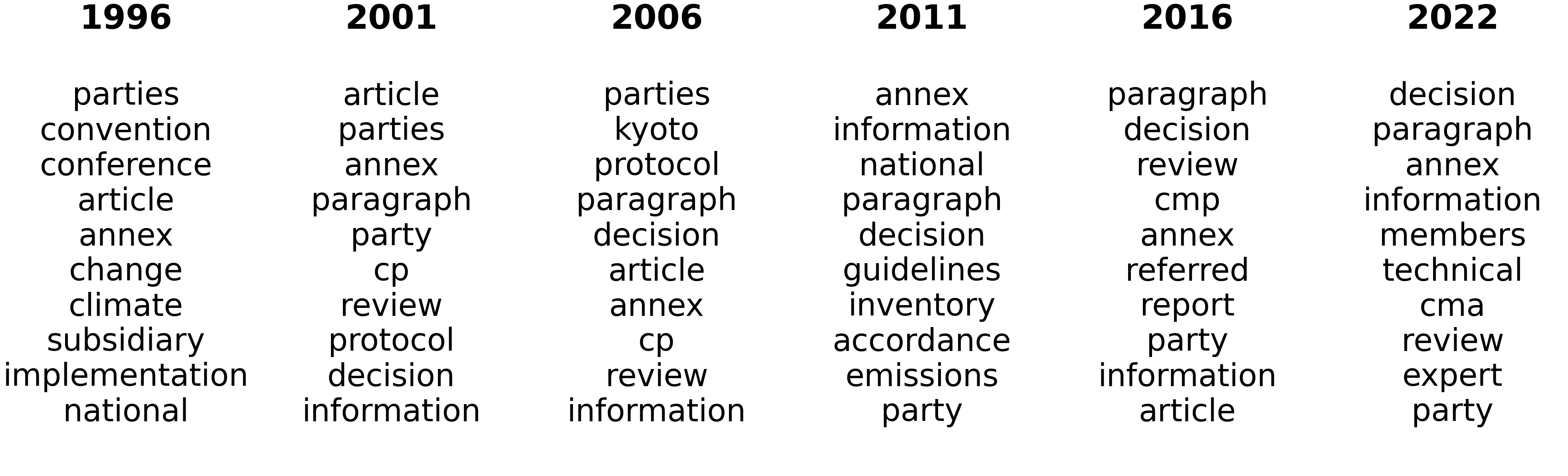}}
    \caption{Temporal evolution of topic 1 ("International Climate Agreements and Frameworks")}
    \label{fig:topic1_evolution}
\end{figure}

\begin{figure}[ht]
    \centering
    \setlength{\fboxsep}{5pt} 
    \setlength{\fboxrule}{2pt} 
    \fbox{\includegraphics[width=0.9\linewidth]{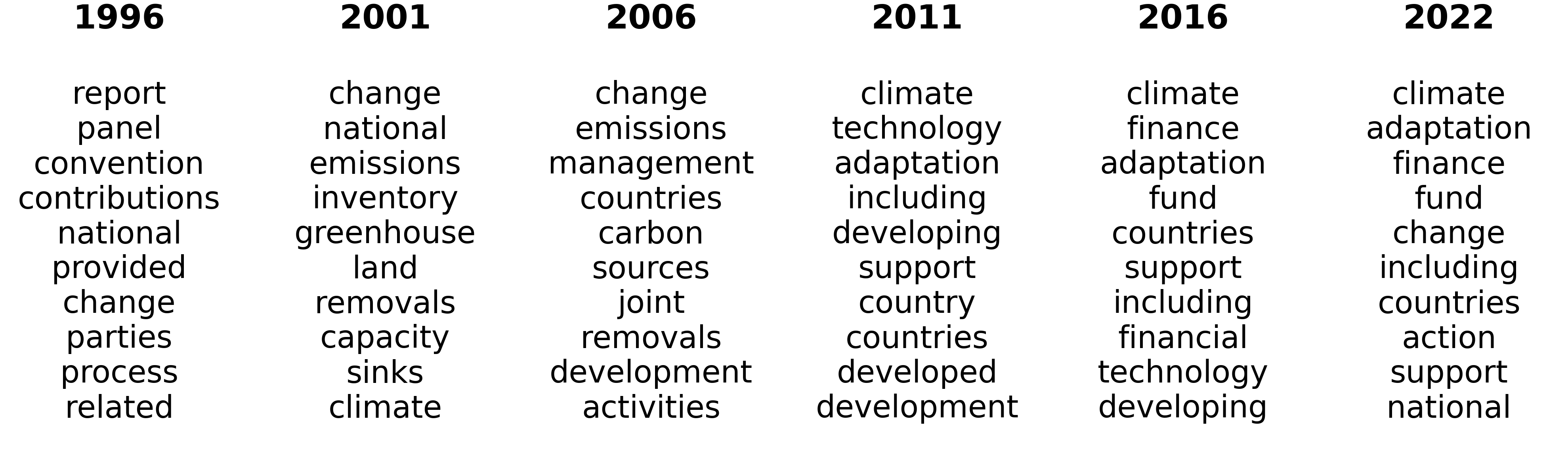}}
    \caption{Temporal evolution of topic 3 ("Conference and Implementation Processes")}
    \label{fig:topic3_evolution}
\end{figure}

\begin{figure}[ht]
    \centering
    \setlength{\fboxsep}{5pt} 
    \setlength{\fboxrule}{2pt} 
    \fbox{\includegraphics[width=0.9\linewidth]{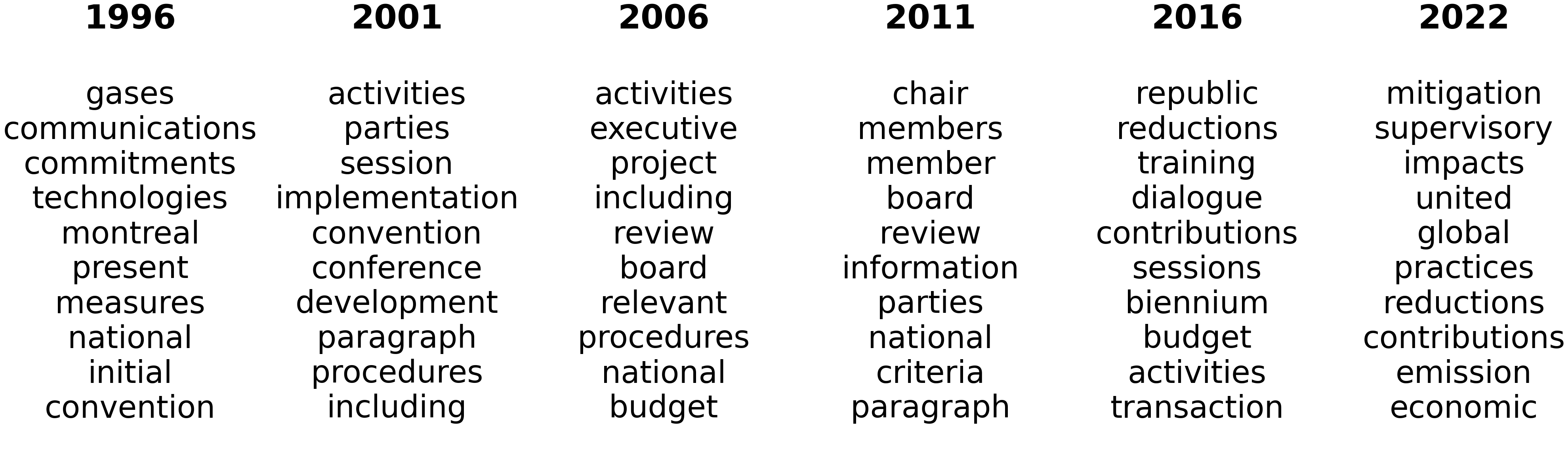}}
    \caption{Temporal evolution of topic 4 ("Technological and Commitment Foundations")}
    \label{fig:topic4_evolution}
\end{figure}

\begin{figure}[ht]
    \centering
    \setlength{\fboxsep}{5pt} 
    \setlength{\fboxrule}{2pt} 
    \fbox{\includegraphics[width=0.9\linewidth]{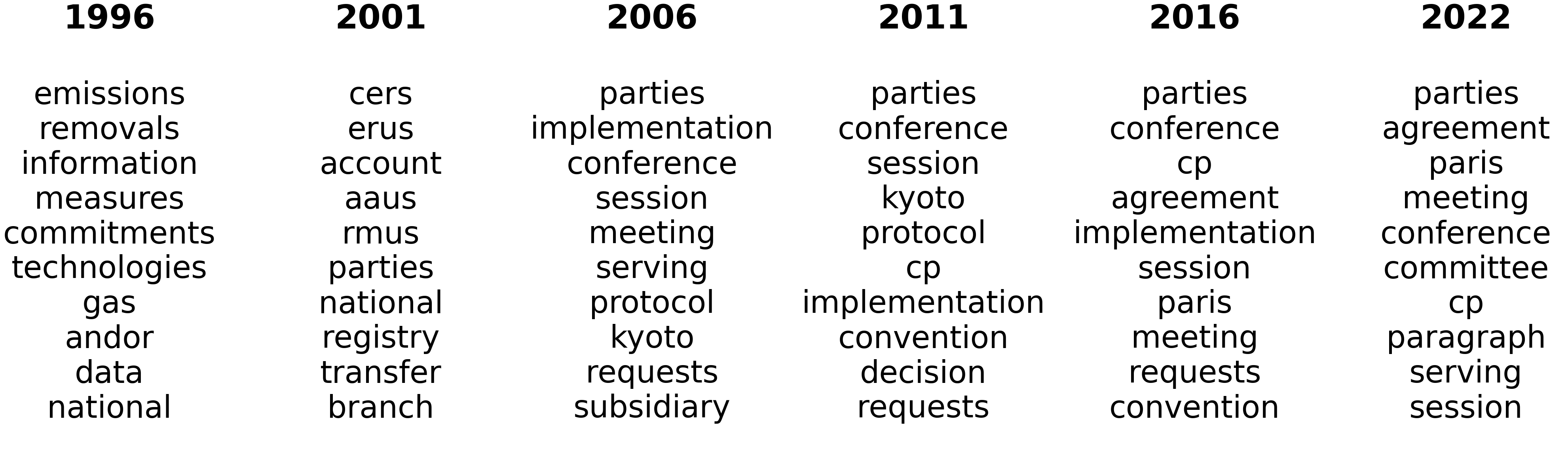}}
    \caption{Temporal evolution of topic 5 ("Emission Management and Reporting")}
    \label{fig:topic5_evolution}
\end{figure}

\newpage

\section{Glossary of Acronyms}
\begin{table}[h!]
\centering
\caption{Acronyms and their full descriptions}
{\renewcommand{\arraystretch}{1.5}
\begin{tabular}{ll}
\hline
\textbf{Acronyms} & \textbf{Description} \\
\hline
UNFCCC & United Nations Framework Convention on Climate Change \\
COP & Conference of the Parties \\
DETM & Dynamic Embedded Topic Model \\
NLP & Natural Language Processing \\
DTM & Dynamic Topic Model \\
LDA & Latent Dirichlet Allocation \\
D-LDA & Dynamic Latent Dirichlet Allocation \\
ETM & Embedded Topic Model \\
CCS & Carbon Capture and Storage \\
GEF & Global Environment Facility \\
WIM & Warsaw International Mechanism \\
WHO & World Health Organization \\
CSV & Comma Separated Values \\
\hline
\end{tabular}
}
\end{table}   

\end{document}